\documentclass[10pt, a4paper]{article}
\usepackage{lrec}
\usepackage{multibib}
\newcites{languageresource}{Language Resources}
\usepackage{graphicx}
\usepackage{tabularx}
\usepackage{soul}
\usepackage{epstopdf}
\usepackage[utf8]{inputenc}

\usepackage{hyperref}
\usepackage{xstring}
\usepackage{verbatim}
\usepackage{color}

\title{CompLex: A New Corpus for Lexical Complexity Prediction from Likert Scale Data}

\name{Matthew Shardlow$^1$, Michael Cooper$^1$, Marcos Zampieri$^2$}

\address{$^1$Manchester Metropolitan University, UK \\
         $^2$Rochester Institute of Technology, USA \\
         M.Shardlow@mmu.ac.uk, mikejcooper90@gmail.com, Marcos.Zampieri@rit.edu}

\abstract{
Predicting which words are considered hard to understand for a given target population is a vital step in many NLP applications such as text simplification. This task is commonly referred to as Complex Word Identification (CWI). With a few exceptions, previous studies have approached the task as a binary classification task in which systems predict a complexity value (complex vs. non-complex) for a set of target words in a text. This choice is motivated by the fact that all CWI datasets compiled so far have been annotated using a binary annotation scheme. Our paper addresses this limitation by presenting the first English dataset for continuous lexical complexity prediction. We use a 5-point Likert scale scheme to annotate complex words in texts from three sources/domains: the Bible, Europarl, and biomedical texts. This resulted in a corpus of 9,476 sentences each annotated by around 7 annotators.
 \\ \newline  }

\begin{document}

\maketitleabstract

\section{Introduction}
 % problem - talk about other datasets
 %  - how do we address this with continuous annotations?
 
 %Whereas previous works have focused on the binary task of Complex word identification (i.e., whether a word is 'complex' or 'not complex'), our focus is on the prediction of lexical complexity (i.e., How complex is a given word). 
 
In many readability applications, it is useful to know the complexity of a given word. In early approaches to the readability task, simple metrics such as whether a word had more than 3 syllables \cite{mc1969smog} or was on a given list or not \cite{dale1948formula} were used to identify complex words. More recently, automated methods for detecting complex words have also been used such as using a threshold on the word's frequency \cite{shardlow:2013:SRW}, or attempting to use a machine learning classifier to determine whether a word is complex or not \cite{CWI2016,yimam2018report}. 
 
These approaches make the fundamental assumption that lexical complexity is binary. That words fall into one of two categories: difficult, or not. Previous approaches to Complex Word Identification (CWI), such as the one used in the CWI shared task (SemEval-2016 Task 11) \cite{CWI2016}, therefore typically refer to binary identification of complex words. A word close to the decision boundary is assumed to be just as complex as one further away. In our work, we move away from this assumption. We theorise that all words are in fact on a continuous scale of complexity and that lexical complexity should be identified accordingly. Binary Complex Word Identification effectively puts an arbitrary threshold on this scale, designating words above or below as complex or simple respectively. In this work, we have foregone the old acronym of CWI, in favour of a new acronym LCP (Lexical Complexity Prediction) which better suits our task of predicting how complex a given word may be. %Marcos: I modified the last sentences.

%Marcos: Maybe we could include an example from the how an instance in the new dataset looks like? 
 
Many factors can be considered to affect lexical complexity prediction. We may consider that the context in which a word is found will affect its understandability. If a word is found in the context of known words, then it may be possible to intuit the meaning from the context. Conversely, a word found in the context of other unknown words may be more difficult to comprehend. Similarly, a reader's familiarity with the genre of the text may affect the perceived complexity of a word. A Biologist reading a Physics journal may struggle with the specialist terms, as would a Physicist reading a Biology journal, but they would each be comfortable with reading material from their own field. 
 
The role of the individual user cannot be overlooked when considering LCP and it is important to consider that although we aim to identify a complexity value for each word, this may need to be adapted for each reader, or group. It may be the case that some words have a high variability (i.e., some readers find them easy and some find them hard), whereas the complexity value of other words is more stable (i.e., all users give the word the same score).
 
Finally, we may wish to consider the effect of multi word expressions on lexical complexity. For example, if I know the complexity value of the constituent words in a multi word expression, can I combine these to give the complexity value of the MWE itself? In some cases, this may be possible (red car is a composition of `red' and `car'), whereas in others it may be more difficult (`European Union' has a deeper meaning than `European' and `Union' combined).
 
In our present work, we introduce CompLex \footnote{\url{https://github.com/MMU-TDMLab/CompLex}}, a new corpus for lexical complexity prediction. we have used crowd sourcing to annotate a new corpus of 8,979 instances covering 3 genres with lexical complexity scores using a 5-point Likert scale (Section~\ref{sec:Dataset}) We have performed baseline experiments to demonstrate the efficacy of a classifier in predicting lexical complexity, as well as further experiments to address some of the open questions as described above (Section~\ref{sec:experiments}) We report our results and discuss our findings throughout (Section~\ref{sec:results})

\section{Related Work}

\subsection{Lexical Complexity}

Given the interest of the community in CWI, two shared tasks on this topic have been organized so far. The first edition of the CWI shared task was the aforementioned SemEval-2016 Task 11 \cite{CWI2016}. In CWI 2016, complexity was defined as whether or not a word is difficult to understand for non-native English speakers. In the CWI 2016 dataset, the annotation followed the binary approach described in the Introduction, where English words in context were tagged as complex or non-complex. The organizers labeled a word as complex in the dataset if the word has been assigned by at least one of the annotators as complex. All words that have not been assigned by at least one annotator as complex have been labeled as non-complex. The task was to use this dataset to train classifiers to predict lexical complexity assigning a label 0 to non-complex words and 1 to complex ones. The dataset made available by the CWI 2016 organizers comprised a training set of 2,237 instances and a much larger test set of 88,221 instances, an unusual setting in most NLP shared tasks where most often the training set is much larger than the test set. 

In \newcite{zampieri2017complex} oracle and ensemble methods have been used to investigate the performance of the participating systems. The study showed that most systems performed poorly due to the way the data was annotated and also due to the fact that lexical complexity was modelled as a binary task, a shortcoming addressed by CompLex.

Finally, a second iteration of the CWI shared task was organized at the BEA workshop 2018 \cite{yimam2018report}. In CWI 2018, a multilingual dataset was made available containing English, German, and Spanish training and testing data for monolingual tracks, and a French test set for multilingual predictions. It featured two sub-tasks: a binary classification task, similar to the CWI 2016 setup, where participants were asked to label the target words in context as complex (1) or simple (0); and a probabilistic classification task where participants were asked to assign the probability that an annotator would find a word complex. The element of regression in the probabilistic classification task was an interesting addition to CWI 2018. However, the continuous complexity value for each word was calculated as the proportion of annotators that found a word complex (i.e., if 5 out of 10 annotators marked a word as complex then the word was given a score of 0.5), a measure which is difficult to interpret as it relies on an aggregation of an arbitrary number of absolute binary judgements of complexity to give a continuous value. 

\subsection{Text Simplification}

Text simplification evaluation is an active area of research, with recent efforts focussing on evaluating the whole process of text simplification in the style of machine translation evaluation. Whilst BLEU score \cite{papineni2002bleu} has been used for text simplification evaluation, this is not necessarily an informative measure, as it inly measures similarity to the target. It does not help a researcher to understand whether the resultant text preserves meaning, or is grammatical. 

To overcome some of these shortcomings, \newcite{xu2016optimizing} introduced the SARI method of evaluating text simplification systems. SARI comprises parallel simplified-unsimplified sentences and measures additions, deletions and those words that are kept by a system. IT does this by comparing input sentences to reference sentences to determine the appropriateness of a simplification. However, SARI is still an automated measure and optimising systems to get a good SARI score may lead to systems that do well on the metric, but not in human evaluations. Recently, EASSE \cite{alva-manchego-etal-2019-easse} has been released to attempt to standardise simplification evaluation by providing a common reference implementation of several text simplification benchmarks.  

Our work does not attempt to simplify a whole sentence through paraphrasing or machine translation, but instead looks at the possibility of identifying which words in a sentence are complex and specifically, how complex those words are. This is a task intrinsically linked to the evaluation of text simplification as the ultimate goal of the task is to reduce the overall complexity of a text. Therefore, by properly understanding and predicting the complexity of words and phrases in a text, we can measure whether it has reduced in complexity after simplification.
 
\section{Dataset} \label{sec:Dataset}
\subsection{Data Collection}
% dataset creation
%  - data gathering
%   - corpora
%   - preprocessing + selelction + frequency band filtering
%   - MWEs

In the first instance, we set about gathering data which we would later annotate with lexical complexity values. We felt it was important to preserve the context in which a word appeared to allow us to understand how the usage of the word affected its complexity. We also allowed multiple instances of each word (up to 5) to allow for cases in our corpus where one word is annotated with different complexity values given different contexts.

\begin{table*}
    \centering
    \begin{tabular}{l|c|c|c|c|c}
                 & Contexts & Unique Words & Median Annotators & Mean Complexity & STD Complexity\\\hline
        All      & 9476  / 7974 / 1500 & 5166 / 3903 / 1263 & 7 / 7 / 7 & 0.394 / 0.385 /  0.442 & 0.110  / 0.108 / 0.105\\
        Europarl & 3496 / 2896 / 600 & 2194  / 1693 / 501 & 7 / 7 / 7.5 & 0.390 / 0.381 / 0.433 & 0.101 / 0.100 / 0.091\\
        Biomed   & 2960 / 2480 / 480 & 1670 / 1250 / 420 & 7 / 7 / 7 & 0.407 / 0.395 / 0.470 & 0.115 / 0.112 / 0.109 \\
        Bible   & 3020 / 2600 / 420  & 1705 / 1362 / 343 & 7  / 7 / 8 & 0.385 /  0.379 / 0.422 & 0.112  / 0.111 / 0.112\\

    \end{tabular}
    \caption{The statistics for CompLex. Each cell shows three values, which are split according to the statistics for `All' / `Single Words' / `Multi Words'}
    \label{tab:stats}
\end{table*}

To add further variation to our data, three corpora were selected as follows:
\begin{description}
 \item[Bible:] We selected the World English Bible translation from \newcite{Christodouloupoulos2015}. This is a modern translation, so does not contain archaic words (thee, thou, etc.), but still contains religious language that may be complex. 
 \item[Europarl:] We used the English portion of the European Pariliament proceedings selected from europarl \cite{koehn2005europarl}. This is a very varied corpus talking about all manner of matters related to european policy. As this is speech transcription, it is often dialogical in nature.
 \item[Biomedical:] We also selected articles from the CRAFT corpus \cite{bada2012concept}, which are all in the biomedical domain. These present a very specialised type of language that will be unfamiliar to non-domain experts.
\end{description}

Each corpus has its own unique language features and styles. Predicting the lexical complexity of diverse sources further distinguishes our work from previous attempts, which have traditionally focused on Wikipedia and News texts.

In addition to single words, we also selected targets containing two tokens  (henceforth referred to as multi word expressions). We used syntactic patterns to identify the multi word expressions, selecting for adjective-noun or noun-noun patterns. We discounted any syntactic pattern that was followed by a further noun to avoid splitting complex noun phrases (e.g., noun-noun-noun, or adjective-noun-noun). Clearly this approach does not capture the full variation of multi word expressions. It limits the length of each expression to 2 tokens and only identifies compound or described nouns. We consider this a positive point as it allows us to make a focused investigation on these common types of MWEs, whilst discounting other less frequent types. The investigation of other types of MWEs may be addressed in a wider study. 

We have not analysed the distribution of compositional vs. non-compositional constructions in our dataset, however we expect both to be present. It would be interesting to further analyse these to distinguish whether the complexity of an MWE can be inferred from tokens in the compositional case, and to what degree this holds for the non-compositional case.

For each corpus we selected words using predetermined frequency bands, ensuring that words in our corpus were distributed across the range of low to high frequency. As frequency is correlated to complexity, this allows us to be certain that our final corpus will have a range of high and low complexity targets. We chose to select 3000 single words and 600 MWEs from each corpus to give a total of 10,800 instances in our pre-annotated corpus. We automatically annotated each sentence with POS tags and only selected nouns as our targets. Again, this limits the field of study, but allows us to make a more focused contribution on the nature of lexical complexity. We have included examples of the contexts, target words and average complexity values in Table \ref{tab:examples}.

\begin{table*}[h]
    \centering
    \begin{tabular}{l|m{8cm}|c}
        \multicolumn{1}{c|}{\textbf{Corpus}} & \multicolumn{1}{c|}{\textbf{Context}} & \textbf{Complexity}  \\ \hline
        Bible & This was the \textbf{length} of Sarah's life. & 0.125 \\
         Biomed & [...] cell \textbf{growth} rates were reported to be 50\% lower [...] & 0.125 \\ 
        Europarl & Could you tell me under which rule they were enabled to extend this item to have four rather than three \textbf{debates}? & 0.208 \\
        Europarl & These agencies have gradually become very important in the \textbf{financial world}, for a variety of reasons.	 & 0.438\\
        Biomed & [...] leads to the \textbf{hallmark loss} of striatal neurons [...] & 0.531 \\
        Bible & The \textbf{idols} of Egypt will tremble at his presence [...] & 0.575 \\
           Bible & This is the law of the \textbf{trespass offering}. & 0.639 \\
          Europarl & They do hold elections, but candidates have to be endorsed by the conservative clergy, so \textbf{dissenters} are by definition excluded.& 0.688 \\
        Biomed & [..] due to a reduction in \textbf{adipose} tissue. & 0.813 \\

    \end{tabular}
    \caption{Examples from out corpus, the target word is highlighted in bold text.}
    \label{tab:examples}
\end{table*}

\subsection{Data Labelling}
%  - data annotation
As has been previously mentioned, prior datasets have focused on either (a) binary complexity or (b) probabilistic complexity. Neither of which give a true representation of the complexity of a word.  In our annotation we chose to annotate each word on  a 5-point Likert scale, where each point was given the following descriptor:
\begin{description}
 \item[1. Very Easy:]  Words which were very familiar to an annotator.
 \item[2. Easy:] Words with which an annotator was aware of the meaning.
 \item[3. Neutral: ]  A word which was neither difficult nor easy.
 \item[4. Difficult:] Words which an annotator was unclear of the meaning, but may have been able to infer the meaning from the sentence.
 \item[5. Very Difficult:] Words that an annotator had never seen before, or were very unclear.
 
\end{description}

We used the following key to transform the numerical labels to a 0-1 range when aggregating the annotations: $1 \rightarrow 0$, $2 \rightarrow 0.25$, $3 \rightarrow 0.5$, $4 \rightarrow 0.75$, $5 \rightarrow 1$. This allowed us to ensure that our complexity labels were normalised in the range 0---1.

%   - crowd sourcing

We employed crowd workers through the figure eight platform, requesting 20 annotations per data instance, paying around 3 cents per annotation. We selected for annotators from English speaking countries (UK, USA and Australia) and selected to disable the use of the Google Translate browser plug-in to ensure that annotators were reading the original source texts and not translated versions of them. In addition, we used the annotation platform's in-built quality control metrics to filter out annotators who failed pre-set test questions, or who answered a set of questions too quickly. 

Our job completed within 3 hours, with over 1500 annotators. The annotators were able to fill in a post-hoc annotation survey, with average satisfaction being around 3 out of 5, the scores typically lower on the `ease of job' metric.

%   - filtering post crowd-sourcing

After we had collected our results, we further analysed the data to detect instances where annotators had not fully participated in the task. We specifically analysed instances where an annotator had given the exact same annotation for all instances (usually these were all 'Neutral') and discarded these from our data. We retained any data instance that had at least 4 valid annotations in our final dataset.

\begin{figure*}
    \centering
    \includegraphics[width=0.88\textwidth]{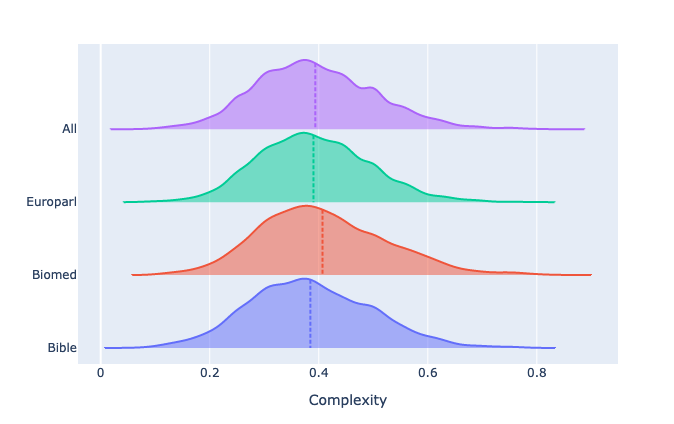}
    \caption{A ridge line plot showing the probability density function of the full dataset (all) as well as each of the genres contained within the full dataset. The vertical dashed line indicates the median in each case.}
    \label{fig:ridge}
\end{figure*}

\subsection{Statistics}

We have provided comprehensive statistics on our corpus in Table \ref{tab:stats}. These show that the average complexity for words in our corpus is 0.395, with a standard deviation of 0.115. A complexity score of 0.5 would be neutral and 0.25 would be easy, so this indicates that on average the words in our corpus fell towards the easier end of the scale. There are however words distributed across the full range of possible complexity annotations as shown by the ridgeline plot in Figure \ref{fig:ridge}. This plot shows the density of complexity annotations in our corpus. It indicates that, whilst the majority of the probability mass is found to the left of the mid-point, there are still many annotations either side of the mid-point for each sub-corpus and for the corpus as a whole.

Table \ref{tab:stats} shows that there was a median of 7 annotators per instance. We requested a total of 20 annotations per instance, but discarded individual annotations that did not meet our inclusion criteria. We discarded any data instances with fewer than 4 annotations. Accordingly, the lowest number of annotations was 4, and the highest  was 20. 

Analysing the sub-genres in our corpus shows some subtle, but meaningful differences between the genres. We used the same inclusion criteria to select words across genres, so as not to bias our results. Bible text and Europarl have very similar average complexity values (0.387 and 0.390), whereas Biomed is higher at 0.407. The biomedical texts are written for a technical audience and can be expected to contain more technical terminology. The bible and europarl may contain genre specific terminology, but will in general reference topics of common knowledge, which will result in higher familiarity and lower complexity. 

We can also see that there is a difference in the complexity level of the annotations between multi word expressions and single words. In the aggregated corpus, single words have an average complexity score of 0.385, whereas multi-words have a higher score of 0.444. This is reflected across each genre, with the largest difference being in biomedical (0.395 / 0.470) and the smallest change being in the Bible (0.380 / 0.428).

\section{Baseline System} \label{sec:experiments}

We developed a baseline for predicting the complexity of a word using our data. We used a linear regression with embedding features for the word and context as well as three hand crafted features, which are known to be strong predictors of lexical complexity. Specifically, the feature sets we used are as follows:

\begin{description}
 \item[Glove Embeddings:] We captured the 300-dimensional Glove embedding \cite{pennington2014glove} for each token in our corpus. This was encoded as 300 separate features (one for each dimension of the embedding).
 \item[InferSent Embeddings:] We captured the 4,096-dimensional embeddings produced by the InferSent library \cite{conneau-EtAl:2017:EMNLP2017} for each context. These were encoded as 4,096 separate features, one for each dimension of the embedding.
 \item[Hand Crafted Features:] We recorded features which are typically known to be strong predictors of lexical complexity. Specifically, we  looked at (1) word  frequency according to the GoogleWeb1T resource \cite{brants2006google}, (2) Word length (as number of characters) and (3) syllable count\footnote{\url{https://pypi.org/project/syllables/}}.
\end{description}

We trained a linear regression using all of these features. We used a held-out test set of 10\% of the data, stratified across corpus type and complexity labels. In addition to this, we also examined the effect of each feature subset. We examined this for the corpus as a whole, as well as for each sub-corpus. These results are presented in Table \ref{tab:results}.

\begin{table}[!ht]
    \centering
    \begin{tabular}{l|c c c c}
             & All    & HC   & Glove   & Sent \\\hline
    All      & 0.1238 & \textbf{0.0853} & 0.0875 & 0.1207 \\
    Bible    & 0.6648 & \textbf{0.0888} & 0.0911 & ---\\
    Biomed   & 0.2954 & \textbf{0.0908} & 0.0939 & ---\\
    Europarl & 0.1982 & \textbf{0.0801} & 0.0879 & ---\\

    \end{tabular}
    \caption{The results of our linear regression  with different feature subsets. We have only reported the sentence embeddings for the whole corpus as the linear regression for the sub-corpora failed to provide a reliable model. All results are reported as mean absolute error. The column headers are as follows: `All' refers to all features concatenated. `HC' refers to hand crafted features, `Glove' refers to the Glove Embeddings (at the target word level) and `Sent' refers to the InferSent embeddings of the contexts.}
    \label{tab:results}
\end{table}

\section{Discussion} \label{sec:results}

Our results show promise for future systems trying to predict lexical complexity by training on continuous data. In the best case, using hand crafted word features such as length, frequency and syllable count, we are able to predict complexity with a mean absolute error of 0.0853. Our values range from 0 (very easy) to 1 (very difficult), so this implies that we would be able to predict complexity with a good degree of accuracy. Features such as length and frequency have long been known to be good predictors of lexical complexity and so it is unsurprising that these ranked highly.

It is interesting to note that the word embeddings performed at a similar level of accuracy (0.0875) to the hand crafted word features. Word embeddings model the context of a word. It may have been the case that certain dimensions of the (300 dimensional) embeddings were more useful for predicting the complexity of a word than others. It would be interesting to further analyse this and to see what contextual information is encoded in the dimensions of these embeddings. It may be that some dimensions encode contexts that rely solely on less frequent, or more frequent words and are therefore better indicators of complexity than others.

Conversely however, the sentence embeddings did not turn out to be good predictors of lexical complexity. These embeddings (4,096 dimensions) were much larger than the word embeddings, which may have made them less suitable for the linear regression. It may be the case that lower dimensional representations of the context would be have more predictive power in our corpus. Although this result implies that context is not important for lexical complexity, we may yet see that future experiments find new ways of integrating the context of the word to better understand it's complexity.

As a classifier, we chose a linear regression. We also used Glove embeddings and infersent. We may find that using embeddings which adapt to the context, such as in BERT and a neural network for prediction would yield stronger results. However, in this work we have only aimed to give an understanding of what types of features can be useful for predicting the values in our corpus, not to produce a state of the art system for the prediction of lexical complexity.

We can see that there are significant differences in the mean absolute error for each sub-corpus. Whereas the mean absolute error was lower for Europarl (0.0801), it was higher for the Bible and Biomed, indicating that the type of language in these two corpora was more difficult to model. This is reflected across different feature subsets, indicating it is a feature of the dataset and not a random fluctuation of our model.

\begin{figure*}
    \centering
    \includegraphics[width =0.70\textwidth]{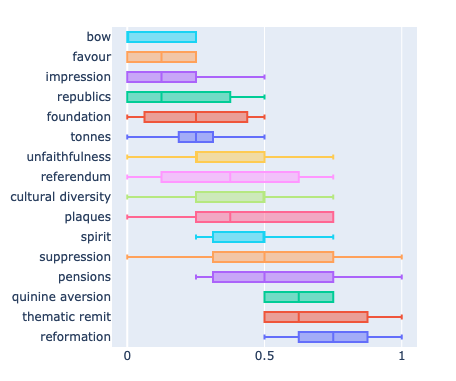}
    \caption{Box plot showing the distribution of annotation scores for different words in CompLex.}
    \label{fig:box}
\end{figure*}

We did not calculate an inter-annotator agreement as part of this work. This is difficult to do in a crowd sourcing setting as we have many annotators and there is no guarantee (or indeed a method to control) whether  the same annotators see a common subset of the annotation data. Instead we used the following principles: (1) We selected for annotators who were known to the platform to provide high quality work. (2) We paid annotators well, encouraging them to take more time over the annotations. (3) We filtered out annotators who had not participated in the task properly. We do not necessarily expect annotators to completely agree on the complexity of a word as one annotator may be  more familiar with a word than another and hence find it easier. We have taken the average values of all annotations for each instance in our corpus, with the hope that this will further smooth out any outliers. In Figure \ref{fig:box}, we have shown a few words and their individual distributions. It is clear that whilst annotators generally agreed on some words, they differed greatly on others. This is reflective of the subjectivity that is present in complexity annotations and warrants further investigation.

\section{Conclusion and Future Work}

In this paper we presented CompLex, a new dataset for lexical complexity prediction. We propose a new 5-point Likert scale annotation scheme to annotate complex words in texts from three sources: the Bible, Europarl, and biomedical texts. The result is a dataset of 9,476 which opens new perspectives in lexical complexity research. We presented multiple baseline experiments using this data and report the best result of 0.0853 mean absolute error. 

Our work leaves many open questions to be answered, and we intend to continue our research to further explore the remaining challenges facing the field of lexical complexity prediction. We have not explored the relationship between the multi-word expressions and single words in our corpus, nor have we explored the transferability of complexity values between corpora. We have also not fully explored the range of classifiers and deep learning architectures that are available from the machine learning community. Again, we expect to cover these in future work. 

\section*{Acknowledgements}

We would like to thank the anonymous reviewers for their valuable feedback. We would also like to thank Richard Evans for insightful discussions on lexical complexity and data annotation. 

\section*{Bibliographical References}\label{reference}
\bibliographystyle{lrec}
\bibliography{lrec}

\begin{thebibliography}{}

\bibitem[\protect\citename{Alva-Manchego \bgroup et al.\egroup
  }2019]{alva-manchego-etal-2019-easse}
Alva-Manchego, F., Martin, L., Scarton, C., and Specia, L.
\newblock (2019).
\newblock {EASSE}: Easier automatic sentence simplification evaluation.
\newblock In {\em Proceedings of EMNLP-IJCNLP}.

\bibitem[\protect\citename{Bada \bgroup et al.\egroup }2012]{bada2012concept}
Bada, M., Eckert, M., Evans, D., Garcia, K., Shipley, K., Sitnikov, D.,
  Baumgartner, W.~A., Cohen, K.~B., Verspoor, K., Blake, J.~A., et~al.
\newblock (2012).
\newblock Concept annotation in the craft corpus.
\newblock {\em BMC bioinformatics}, 13(1):161.

\bibitem[\protect\citename{Brants and Franz}2006]{brants2006google}
Brants, T. and Franz, A.
\newblock (2006).
\newblock The google web 1t 5-gram corpus version 1.1.
\newblock {\em LDC2006T13}.

\bibitem[\protect\citename{Christodouloupoulos and
  Steedman}2015]{Christodouloupoulos2015}
Christodouloupoulos, C. and Steedman, M.
\newblock (2015).
\newblock A massively parallel corpus: the bible in 100 languages.
\newblock {\em Language Resources and Evaluation}, 49(2):375--395, Jun.

\bibitem[\protect\citename{Conneau \bgroup et al.\egroup
  }2017]{conneau-EtAl:2017:EMNLP2017}
Conneau, A., Kiela, D., Schwenk, H., Barrault, L., and Bordes, A.
\newblock (2017).
\newblock Supervised learning of universal sentence representations from
  natural language inference data.
\newblock In {\em Proceedings of EMNLP}.

\bibitem[\protect\citename{Dale and Chall}1948]{dale1948formula}
Dale, E. and Chall, J.~S.
\newblock (1948).
\newblock A formula for predicting readability: Instructions.
\newblock {\em Educational research bulletin}, pages 37--54.

\bibitem[\protect\citename{Koehn}2005]{koehn2005europarl}
Koehn, P.
\newblock (2005).
\newblock Europarl: A parallel corpus for statistical machine translation.
\newblock In {\em Proceedings of MT Summit}.

\bibitem[\protect\citename{Mc~Laughlin}1969]{mc1969smog}
Mc~Laughlin, G.~H.
\newblock (1969).
\newblock Smog grading-a new readability formula.
\newblock {\em Journal of reading}, 12(8):639--646.

\bibitem[\protect\citename{Paetzold and Specia}2016]{CWI2016}
Paetzold, G.~H. and Specia, L.
\newblock (2016).
\newblock {SemEval 2016 Task 11: Complex Word Identification}.
\newblock In {\em Proceedings of SemEval}.

\bibitem[\protect\citename{Papineni \bgroup et al.\egroup
  }2002]{papineni2002bleu}
Papineni, K., Roukos, S., Ward, T., and Zhu, W.-J.
\newblock (2002).
\newblock Bleu: a method for automatic evaluation of machine translation.
\newblock In {\em Proceedings of ACL}.

\bibitem[\protect\citename{Pennington \bgroup et al.\egroup
  }2014]{pennington2014glove}
Pennington, J., Socher, R., and Manning, C.~D.
\newblock (2014).
\newblock Glove: Global vectors for word representation.
\newblock In {\em Proceedings of EMNLP}.

\bibitem[\protect\citename{Shardlow}2013]{shardlow:2013:SRW}
Shardlow, M.
\newblock (2013).
\newblock {A Comparison of Techniques to Automatically Identify Complex Words}.
\newblock In {\em Proceedings of the ACL Student Research Workshop}.

\bibitem[\protect\citename{Xu \bgroup et al.\egroup }2016]{xu2016optimizing}
Xu, W., Napoles, C., Pavlick, E., Chen, Q., and Callison-Burch, C.
\newblock (2016).
\newblock Optimizing statistical machine translation for text simplification.
\newblock {\em Transactions of the Association for Computational Linguistics},
  4:401--415.

\bibitem[\protect\citename{Yimam \bgroup et al.\egroup }2018]{yimam2018report}
Yimam, S.~M., Biemann, C., Malmasi, S., Paetzold, G., Specia, L.,
  {\v{S}}tajner, S., Tack, A., and Zampieri, M.
\newblock (2018).
\newblock A report on the complex word identification shared task 2018.
\newblock In {\em Proceedings of BEA}.

\bibitem[\protect\citename{Zampieri \bgroup et al.\egroup
  }2017]{zampieri2017complex}
Zampieri, M., Malmasi, S., Paetzold, G., and Specia, L.
\newblock (2017).
\newblock Complex word identification: Challenges in data annotation and system
  performance.
\newblock In {\em Proceedings of NLP-TEA}.

\end{thebibliography}

\end{document}